\documentclass[tinyml]{acmart}
\acmSubmissionID{11}
\usepackage{graphicx}
\usepackage{tikz}
\usepackage{caption}
\usepackage{subcaption}
\usepackage{multirow}
\usepackage{chngcntr}
\usepackage{url}

\AtBeginDocument{%
  \providecommand\BibTeX{{%
    \normalfont B\kern-0.5em{\scshape i\kern-0.25em b}\kern-0.8em\TeX}}}

\setcopyright{rightsretained}
\copyrightyear{2022}
\acmYear{2022}

\begin{document}

\title{An Empirical Study of Low Precision Quantization for TinyML}


\author{\small Shaojie Zhuo}
\authornote{Equal contribution}
\affiliation{\footnotesize%
  \institution{Qualcomm AI Research}
  \institution{Qualcomm Canada ULC}
}
\email{shaojiez@qti.qualcomm.com}

\author{\small Hongyu Chen}
\authornotemark[1]
\affiliation{\footnotesize%
  \institution{University of Toronto}
}
\email{hy.chen@mail.utoronto.ca}

\author{\small Ramchalam Kinattinkara Ramakrishnan}
\affiliation{\footnotesize %
  \institution{Qualcomm AI Research}
  \institution{Qualcomm Canada ULC}
}
\email{rkinatti@qti.qualcomm.com}

\author{\small Tommy Chen}
\affiliation{\footnotesize %
  \institution{Qualcomm AI Research}
  \institution{Qualcomm Canada ULC}
}
\email{tommchen@qti.qualcomm.com}

\author{\small Chen Feng}
\affiliation{\footnotesize %
  \institution{Qualcomm AI Research}
  \institution{Qualcomm Canada ULC}
}
\email{chenf@qti.qualcomm.com}

\author{\small Yicheng Lin}
\affiliation{\footnotesize %
  \institution{Qualcomm AI Research}
  \institution{Qualcomm Canada ULC}
}
\email{yichengl@qti.qualcomm.com}

\author{\small Parker Zhang}
\affiliation{\footnotesize %
  \institution{Qualcomm AI Research}
  \institution{Qualcomm Canada ULC}
}
\email{xiaopeng@qti.qualcomm.com}

\author{\small Liang Shen}
\affiliation{\footnotesize %
  \institution{Qualcomm AI Research}
  \institution{Qualcomm Canada ULC}
}
\email{liangs@qti.qualcomm.com}


\renewcommand{\shortauthors}{Authors, et al.}

\begin{abstract}
Tiny machine learning (tinyML) has emerged during the past few years aiming to deploy machine learning models to embedded AI processors with highly constrained memory and computation capacity. Low precision quantization is an important model compression technique that can greatly reduce both memory consumption and computation cost of model inference. In this study, we focus on post-training quantization (PTQ) algorithms that quantize a model to low-bit (less than 8-bit) precision with only a small set of calibration data and benchmark them on different tinyML use cases. To achieve a fair comparison, we build a simulated quantization framework to investigate recent PTQ algorithms. Furthermore, we break down those algorithms into essential components and re-assembled a generic PTQ pipeline. With ablation study on different alternatives of components in the pipeline, we reveal key design choices when performing low precision quantization. We hope this work could provide useful data points and shed lights on the future research of low precision quantization.
\end{abstract}


\maketitle

\section{Introduction}
\label{sec:intro}
Embedding AI directly on edge devices is becoming a key to revolutionize the internet of things (IoT), where billions of tiny devices are leveraged to gain great productivity and efficiency in areas including consumer, medical, automotive and industrial. As a result, tiny machine learning (tinyML) has emerged over the past few years that aims to deploy machine learning algorithms, especially neural networks, on embedded AI processors such as micro-controllers or embedded NPUs (e.g. Qualcomm's eNPU~\cite{enpu2021} and ARM's MicroNPU~\cite{micronpu2021}) with very low power consumption in the level of a few milli-watts. By performing inferences on the devices, tinyML brings the advantage of high energy efficiency, fast responsiveness, high privacy and strong autonomy of edge devices. However, the embedded AI processors are usually highly resource-constrained with limited memory and computation capability. A large body of literature has focused on addressing these issues by making neural networks more efficient, including efficient neural architecture search~\cite{zoph2016, liu2019, cai2020}, neural networks and system co-design~\cite{lin2020mcunet, lin2021mcunetv2} and model compression such as pruning~\cite{han2016}, knowledge distillation~\cite{hinton2015} and quantization~\cite{han2016, krishnamoorthi2018}.

Neural network quantization is one of the most important model compression techniques. It compresses models by using low-bit precision representation for weight and activation tensors, instead of the 32-bit (or 16-bit for half-precision) precision that is commonly used during model training. The memory consumption and computation of a low precision model are greatly reduced. And quantization compresses a model without changing its architecture. This is particularly useful when the model architecture is already optimally designed for a specific inference system or device. Besides, quantization can be applied along with other model compression techniques, such as pruning and knowledge distillation.  

Neural networks have been shown to be quite robust to quantization with 8-bit precision without sacrificing the accuracy\cite{nagel2021}. Quantization with lower precision can further reduce the memory consumption and computation. And ultra-low precision (1 or 2-bit) operations can often be computed efficiently with bit-wise arithmetic and thus achieving signification computation acceleration~\cite{zhu2022tab}. However, due to the large quantization noise, the benefits of low precision quantization often come at the cost of significant accuracy degradation. Among approaches to mitigate the accuracy drop caused by quantization, post-training quantization (PTQ) directly performs quantization on a pre-trained full precision model to convert it to a corresponding low precision model. Unlike quantization-aware training (QAT), PTQ does not require the full training pipeline of the original model and can be applied with a small amount of calibration data (or even data-free~\cite{nagel2019}), thus becoming a popular model 
quantization routine. 

In this paper, we aim to study and benchmark recent PTQ algorithms on neural network models designed for tinyML use cases. By doing that, we try to answer the following questions.

First of all, \textit{what is the performance of PTQ algorithms on tinyML models?} Existing work~\cite{nagel2019, nagel2020adaround, li2021brecq, esser2020lsq, wang2020} has shown successful applications of low precision PTQ on big neural network models for large-scale tasks such as image classification on ImageNet. However, tinyML models are usually very compact and are designed for relatively simple tasks. It is unknown whether those models can be quantized to low precision while preserving their accuracy. 

Secondly, \textit{what are the design choices making a PTQ algorithm superior to others?} Comparing the performance of different PTQ algorithms helps us to select a proper quantization algorithm. And further study on their algorithm design choices can reveal the key reason of an algorithm's success and provide insights for further improvement of the algorithm. 

Last but not least, \textit{what is the trade-off between accuracy and memory/computation when applying low precision quantization?} This is an important question when deploying a tinyML model. By knowing the sensitivity of different models to quantization, we are able to seek the best trade-off between accuracy and memory/computation considering the very limited resources on tinyML devices.

By conducting extensive experiments with various tinyML models, we find answers to the previous questions and many insightful observations. To summarize our contributions:

\begin{itemize}
    \item We create a quantization simulation framework and an unified PTQ pipeline to encourage fair comparison of different PTQ algorithms.
    \item We conduct ablations to reveal the key design choices of the PTQ pipeline using a set of models with varieties of architectures in tinyML.
    \item we demonstrate that low precision quantization is useful for tinyML by compressing a model to greatly save memory and computations while preserving accuracy.
    \item We find key gaps of low precision quantization for tinyML and point out the directions of potential improvements. 
\end{itemize}

We believe that low precision quantization is important to tinyML. And the observations, challenges and open questions in this study are good data points to understand the existing PTQ algorithms and the characteristic of low precision quantization, and shed lights on the direction of future work to improve the performance.

\section{Related work}
\label{sec:relatedwork}
\paragraph{Quantization} We refer to the readers a good overview of quantization basics and algorithms in~\cite{gholami2021survey}. Quantization algorithms can be classified into quantization-aware training (QAT) and post-training quantization (PTQ). QAT re-trains a model with quantized parameters so that it converges to a point with a better loss. An important challenge of the backpropagation during re-training is how to treat the non-differentiable rounding operator in the quantization function. Straight Through Estimator (STE)~\cite{bengio2013ste} is one of the main techniques used to approximate the gradient. Built on top of that, several recent papers~\cite{esser2020lsq,jain2020tqt} proposed to learn the quantization parameters together with the network parameters and achieve better results. Although STE-based methods enable the training of quantized networks with gradient-based optimization, the gradient mismatch between forward and backward passes affects the training effectiveness, especially for low precision quantization. To alleviate the gradient mismatch, soft quantizers with sigmoid or tanh function~\cite{gong2019dsq} are used to approximate the discrete rounding in both forward and backward passes. These approaches, however, cause the accuracy drop due to the discrepancy between the soft rounding during training and the hard rounding at inference time. Despite the promising results given by QAT methods, they usually need the whole training pipeline and much more computation. In contrast, PTQ is attractive in the sense that it requires only a small set of (unlabeled) calibration data and thus become the main focus of this study. A fundamental problem in PTQ is finding good quantization parameters for both weight and activation tensors. It is particular challenging when there is imbalanced distribution among different channels of a tensor. To resolve that, cross layer equalization (CLE)~\cite{nagel2019} and outlier channel splitting (OCS)~\cite{zhao2019} are proposed to re-balance the distribution of weights. Although good performance are achieved for 8-bit quantization, PTQ suffers from severe performance degradation as precision goes lower than 8-bit. Recently, a new paradigm makes huge progress for low precision PTQ. The approach utilizes layerwise optimization to minimize the error between the output of a quantized layer and that of its corresponding full-precision layer, with the help of a small set of calibration data. A series of work leveraging the layer-wise calibration paradigm, namely AdaRound~\cite{nagel2020adaround}, AdaQuant~\cite{hubara2020adaquant}, BitSplit~\cite{wang2020}, BRECQ~\cite{li2021brecq}, has push the limit of PTQ down to 2 bit. We thus focus on the study of those methods and compare their performances.

\paragraph{Quantization for tinyML} There are works~\cite{krishnamoorthi2018, wu2020integer} focusing on the study of quantization for general applications with big models in the study. In the context of tinyML,~\cite{Pierre2021} proposed a new QAT method for low precision quantization.~\cite{ghamari2021, banbury2021} focused on deployment of the 8-bit quantized models and their latency on actual tinyML devices. By contrast, Our study aims on benchmarking the performance of the state of the art quantization methods on low precision (less than 8-bit) quantization for tinyML. We selected several representative applications in tinyML to study the effects of different algorithm design choices and the trade-off among accuracy, memory consumption and computational cost.

\section{Quantization framework}
\label{sec:framework}
The performances of existing PTQ algorithms are usually reported based on separate implementations without a common framework. The  implementation differences (e.g. the way how quantization is simulated) makes it difficult to do a fair comparison of different algorithms. And existing work usually focuses on improvement of different aspects in the quantization process without optimizing the whole pipeline. To this end, we implement a quantization framework to facilitate the quantization study. Built on top of that, we propose a unified post-training quantization pipeline, so that existing PTQ algorithms can be deconstructed and fitted into the pipeline as improvement of parts of its components. It does not only make fair comparisons of PTQ algorithms possible, but also has the flexibility to ablate on different choices of each components in the pipeline and find the optimal one that generate the best performance. In this section, we introduce our quantization simulation framework and the proposed unified PTQ pipeline.

\begin{figure}[htb]
  \centering
  \includegraphics[trim=1cm 0.5cm 1cm 1cm, width=0.15\textwidth]{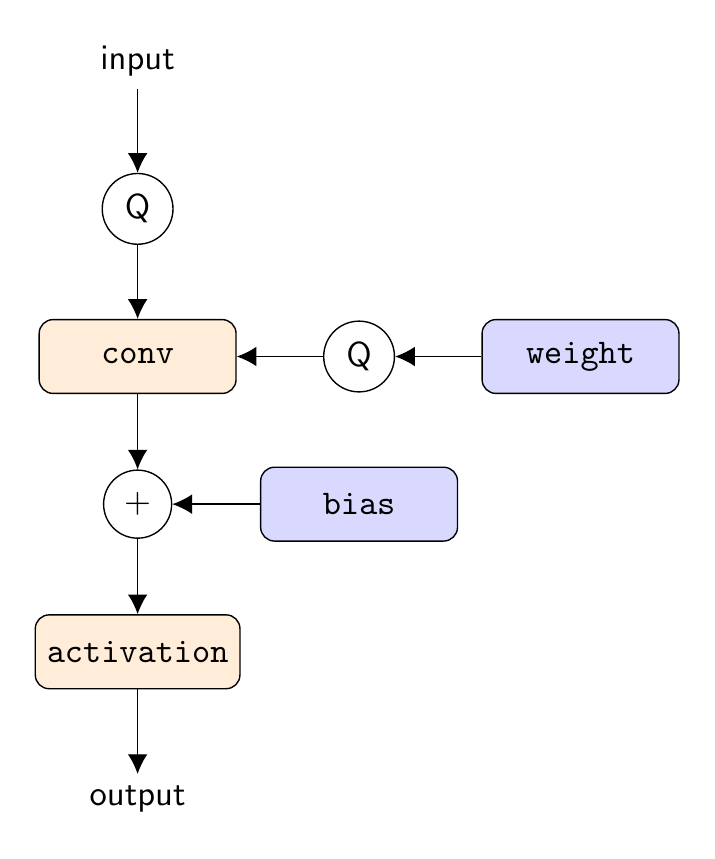}
  \caption{Simulated quantization for the forward pass of a convolution layer.}
  \label{fig.sim_quant}
\end{figure} 
\subsection{Simulated quantization}
We use simulated quantization~\cite{krishnamoorthi2018} in our framework. A tensor is quantized and then de-quantized back to floating precision to simulate quantization loss while the operations are carried out with floating point arithmetic. Figure~\ref{fig.sim_quant} shows an example of the quantized forward pass of a convolution layer. Both input and weight tensor go through the quantization function in the quantizer to simulate quantization and then are fed to the convolution computation in full precision. The bias remains in full precision. The quantization of other operators is simulated in a similar way. 

\begin{figure*}[!htb]
  \centering
  \includegraphics[trim=1cm 0.5cm 1cm 0.5cm,width=0.6\textwidth]{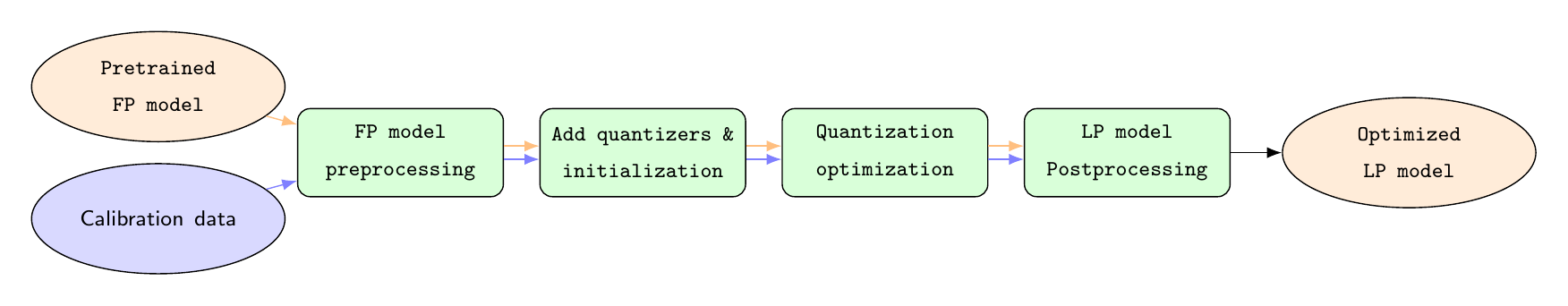}
  \caption{The unified pipeline for PTQ. It takes a pre-trained full precision model and a small set of calibration data as input and output a optimized quantized model through multiple steps.}
  \label{fig.pipeline}
\end{figure*}
\begin{figure*}
     \centering
     \begin{subfigure}[b]{0.3\textwidth}
         \centering
         \includegraphics[trim=1cm 0.5cm 1cm 0.5cm, width=0.3\textwidth]{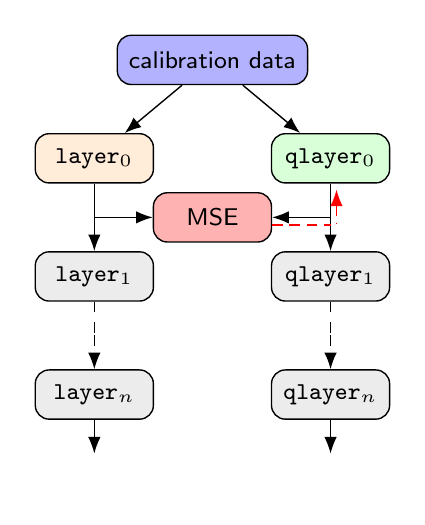}
         \caption{optimization of the first quantized layer}
         \label{fig:s1}
     \end{subfigure}
     \hfill
     \begin{subfigure}[b]{0.3\textwidth}
         \centering
         \includegraphics[trim=1cm 0.5cm 1cm 0.5cm, width=0.3\textwidth]{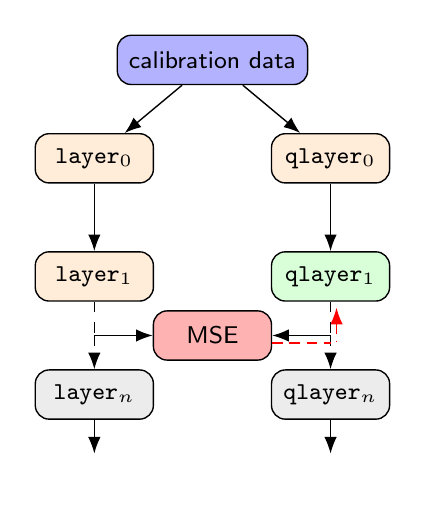}
         \caption{optimization of the second quantized layer}
         \label{fig:s2}
     \end{subfigure}
     \hfill
     \begin{subfigure}[b]{0.3\textwidth}
         \centering
         \includegraphics[trim=1cm 0.5cm 1cm 0.5cm, width=0.3\textwidth]{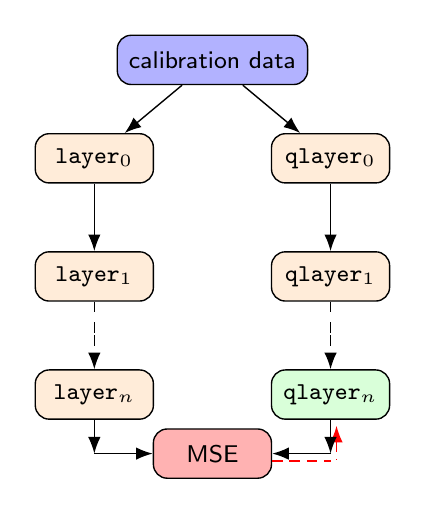}
         \caption{optimization of the last quantized layer}
         \label{fig:sn}
     \end{subfigure}
     \hfill
        \caption{Layerwise quantization optimization. With a small set of calibration data, (a) the first quantized layer (in \textcolor{green!50}{green}) is activated and optimized, while the rest of the quantized layers stay inactive (in \textcolor{gray!50}{gray}). The first quantizated is optimized to minimize the mean squared error(MSE) between its output and that of the first FP layer; (b) the first quantized layer is optimized and frozen (in \textcolor{orange!50}{orange}) and the second quantized layer is activated and optimized; (c) The quantized model is optimized layer by layer until the last one.}
        \label{fig:layer_opt}
\end{figure*}

We adopt the asymmetric quantization scheme in our framework for quantization simulation. The asymmetric quantization function is defined by three quantization parameters: the scale factor $s$, the zero-point $z$ and the bit-width $b$, 

\begin{equation}
\label{equ:quantization}
    \hat{x} = q(x;s,z,b)=s\Big[clamp \Big(\Big\lfloor \frac{x}{s} \Big\rceil+z, n, p\Big)-z\Big]
\end{equation}
where $\hat{x}$ is the quantized approximation of $x$, $\lfloor\cdot\rceil$ indicates the round function and $clamp(\cdot)$ clamps values between $n$ and $p$, which are determined by the bit-width $b$ and define the range of the integer grids. As an simplified version of the asymmetric quantization, symmetric quantization restricts the zero-point to be $0$ to reduce the additional computation introduced by the zero-point offset. We follow~\cite{nagel2021} to use \textit{asymmetric quantization for activation tensors and symmetric quantization for parameters}, considering that the distributions of activation tensors are often asymmetric about zero while those for parameters are roughly symmetric about zero. 

Quantization granularity is another factor affecting the accuracy of a quantized model. Per-tensor quantization has a single set of quantization parameter for the whole tensor while per-channel quantization has a separate set of quantization parameters for each channel of the tensor. Per-channel quantization usually introduces lower quantization noise due to its finer granularity. In our framework, we use \textit{per-channel quantization for parameters and per-tensor quantization for activation tensors}, considering the readiness of hardware support and to better preserve the accuracy of a model~\cite{nagel2021}.

\subsection{PTQ Pipeline}
\label{sec.ptq_pipeline}
As illustrated in Figure \ref{fig.pipeline}, the proposed unified PTQ pipeline takes a pre-trained full precision model with a set of calibration data as input and produces the optimized low precision model through four sequential steps.

\textit{Step 1: FP model preprocessing.} In this step, a full precision (FP) model is pre-processed to make it more quantization friendly. One common issue for quantization is the imbalanced distribution among various channels of the tensor, which makes it hard to find a suitable quantization parameters for the tensor. This is especially prevalent in the weights of efficient models with depth-wise separable convolutions~\cite{nagel2019}. Fine-grained quantization (e.g., per-channel quantization) is able to alleviate the issue. Without it, cross-layer range equalization (CLE)~\cite{nagel2019} and Outlier channel splitting (OCS)~\cite{zhao2019} are introduced to re-balance the distribution of a tensor. Although we use per-channel quantization for parameters, the activation tensors are quantized using per-tensor quantization. It is worthwhile to show the impact of FP model preprocessing on quantization.

\textit{Step 2: Add quantizers and initialization.} After the preprocessing, we add quantizers to a model to simulate quantization for both parameters and activation tensors. To initialize the quantization parameters (the scale $s$ and zero-point $z$) of each quantizer, one common approach is to find the clipping range $[\alpha, \beta]$ first and then use it to calculate the scale and zero-point, i.e., 
\begin{equation}
s=\frac{\beta-\alpha}{2^b-1}, z=\lfloor\frac{-\alpha}{s}\rceil+n
\end{equation}
And there are multiple ways to determine the clipping range,
\begin{itemize}
    \item \textit{MinMax} initialization sets the clipping range using the min/max of a tensor. It is simple but usually suffers from outliers.
    \item \textit{MSE} initialization finds the range that minimize the mean squared error (MSE) between the original and quantized tensor using grid search or analytical approximations $\alpha^{*},\beta^{*}=\mathop{\arg \min}_{\alpha, \beta}\|x-\hat{x}\|^2_F $,where $\hat{x}$ is the quantized version of $x$ and $\|\cdot\|_F$ is the Frobenius norm. 
\end{itemize}

\textit{Step 3: Quantization optimization.} As shown in Equation \ref{equ:quantization}, the quantization function maps a floating point value to a low-precision value and then de-quantize it back to a floating point value. During the process, clipping error is introduced by the $clamp(\cdot)$ function and rounding error is introduced by the round function $\lfloor\cdot\rceil$. Quantization optimization is the process to to seek the right trade-off between the two errors. We adopt a general layerwise optimization process, as illustrated in Figure~\ref{fig:layer_opt}. A low precision model is optimized layer by layer to minimized the loss (usually the mean squared error) between the outputs of the full precision layer and the quantized layer. As shown in equation~\ref{equ:quantization}, multiple variables can potentially be optimized,
\begin{itemize}
    \item Optimizing scale and zero-point (\textit{opt\_qparam}). Learnt step-size quantization~\cite{esser2020lsq} treats the scale $s$ and zero-point $z$ as trainable parameter and optimizes them to minimize the layerwise MSE.
    \item Optimizing weights (\textit{opt\_weights}). AdaQuant~\cite{hubara2020adaquant} extends the flexibility and optimizes the weights directly to minimize the layerwise MSE. 
    \item Optimizing bits (\textit{opt\_bits}). Instead of working on the floating point, BitSplit~\cite{wang2020} performs the optimization on the quantized values directly and splits the integer into multiple bits and optimizes them to minimize the layerwise MSE. 
    \item Optimizing rounding (\textit{opt\_round}). AdaRound~\cite{nagel2020adaround} optimizes the rounding by introducing a auxiliary parameter for each weight and constraints it to be either $0$ or $1$ to achieve rounding up or down.
\end{itemize}
In the qunatization optimization step, instead of layer-wise optimization, block-wise optimization~\cite{li2021brecq} is also proposed to capture the cross-layer dependency within the block and it is clained that block-wise optimization could be a better bias-variance trade-off choice with limited calibration data for low precision quantization. 

\textit{Step 4: LP model post-processing.} After quantization optimization, bias tuning~\cite{hubara2020adaquant} can be applied on the low-precision (LP) model to update biases in a model to further compensate for quantization loss. The biases are updated to minimize the MSE of the final outputs between FP model and the LP model. In theory, the bias tuning can be done during the quantization optimization, but the rationale to put it in the post-processing stage is that, by updating a limited part of the model parameters in a end-to-end fashion, it can provide global guidance to optimize for the final output without over-fitting.

With the unified pipeline, existing PTQ algorithms can be deconstructed and fitted into the pipeline. We vary each component in the pipeline to construct new PTQ algorithms. We do ablations on different options for each component to reveal the best options and then use them to construct the best PTQ algorithm.

\section{Experiments}
\label{sec:exp}
In this section, we benchmark and ablate the performance of recent PTQ algorithms on multiple tinyML models. We first describe the tinyML tasks and models selected for the experiments and then show the experiment results of ablations on different components of the PTQ pipeline and results of end-to-end study. In our experiments, for all the models, we report the mean and standard deviation of the accuracy, calculated using 5 runs with different initial seeds. And we use 1024 unlabeled data extracted from the original training data as the calibration dataset.

\subsection{TinyML tasks and models}
We select four use cases for our study: image classification, visual wake words, Keyword spotting and human activity recognition. The first three are from the MLPerf Tiny benchmark~\cite{banbury2021}. The selection is based on the relevance to real world applications, the availability of open-sourced dataset and models, and the coverage of a wide variety of applications and a wide variety of model architectures in tinyML. We train each model based on the recommended setting in~\cite{banbury2021} to get the full precision models. A summary of the tasks and models are shown in Table \ref{tab:usecase}.

\begin{table*}[htb]
\centering
\begin{scriptsize}
\begin{tabular}{|c|c|c|c|c|c|c|}
\hline
\textbf{Use case} & \textbf{Dataset} & \textbf{Model} & \textbf{Float Accuracy} & \textbf{MAC} &\textbf{Num Param.} & \textbf{Peak activation}\\ \hline
Image Classification & CIFAR10 & Res8 & $86.75$ & $12,591,808$ & $77,706$ & $36,608$\\ \hline
Keyword Spotting & Speech Commands & DS-CNN & $93.03$ & $2,736,832$ & $22,604$ & $36,864$ \\ \hline
Visual Wake Words & VWW & MobileNetV1 & $85.10$ & $7,723,776$ & $210,850$ & $32,768$\\ \hline
Human Activity Recognition & UCI-HAR & CNN & $90.36$ & $2,298,368$ & $523,462$ & $8,064$ \\ \hline
\end{tabular}
\end{scriptsize}
\caption{TinyML tasks and models selected for the evaluation of quantization algorithms. The detailed architecture of each model can be found in the appendix}
\label{tab:usecase}
\end{table*}

\subsection{Ablations of the PTQ pipeline}
\begin{figure}[htb]
  \centering
  \begin{tabular}{cc}
    \subcaptionbox{\label{1st-fig}Initialization: Res8 (CIFAR10)}{\includegraphics[trim=2cm 0.5cm 2cm 0.5cm,width=0.46\linewidth]{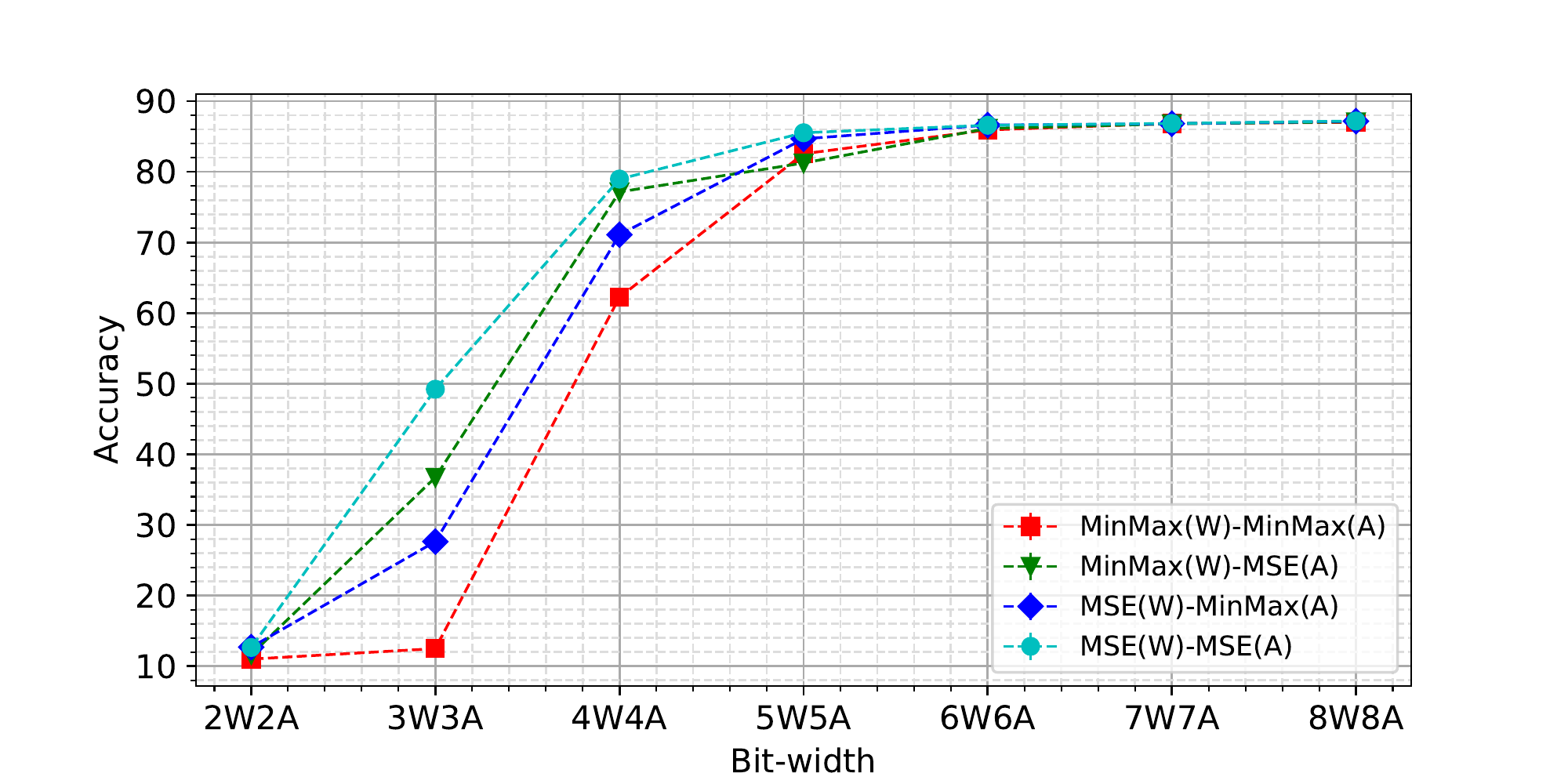}} 
    & 
    \subcaptionbox{\label{2st-fig}Initialization: DS-CNN(Speech Command)}{\includegraphics[trim=2cm 0.5cm 2cm 0.5cm,width=0.46\linewidth]{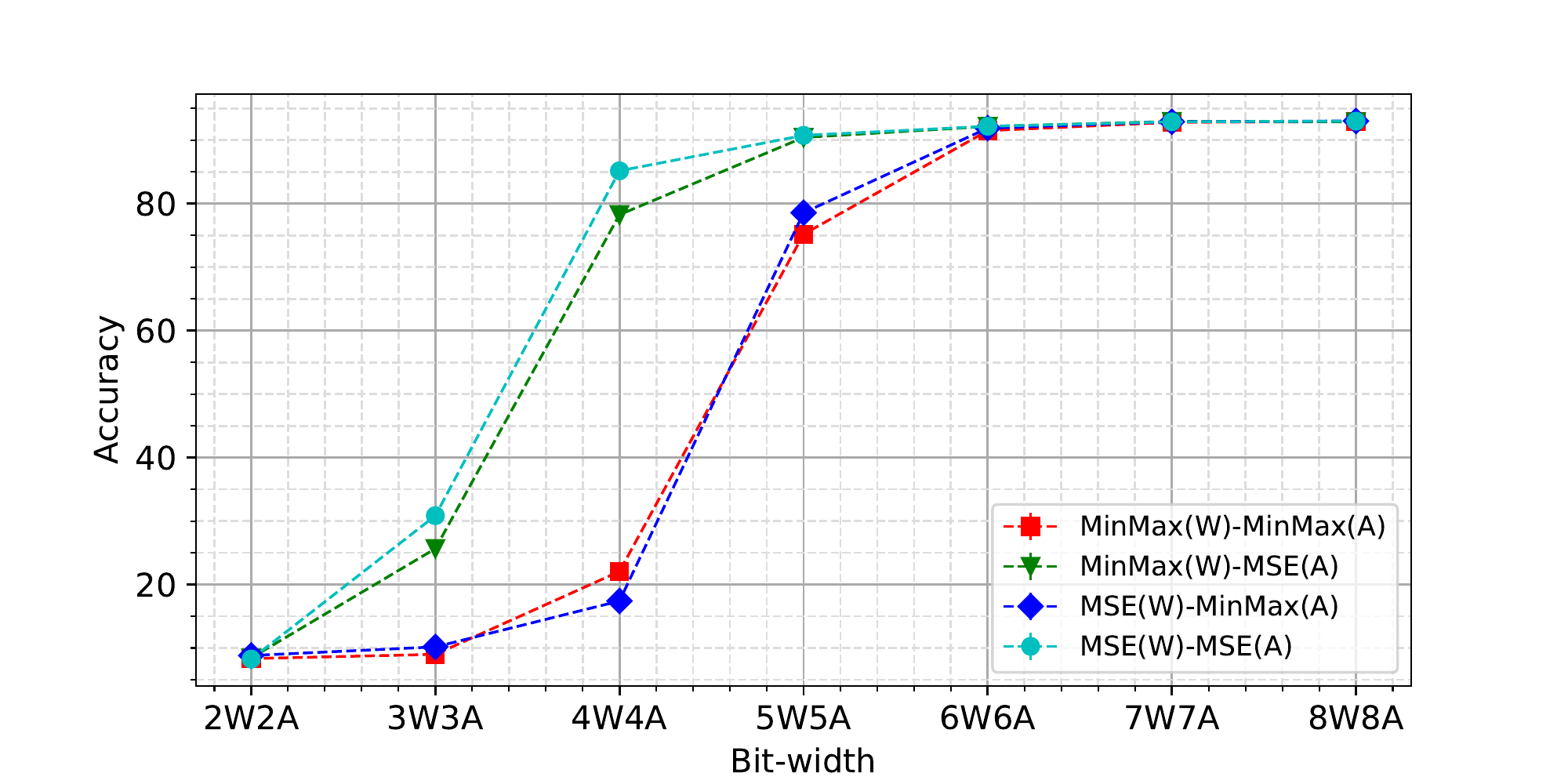}} 
    \\
    \subcaptionbox{\label{1st-fig}Pre-processing: Res8 (CIFAR10)}{\includegraphics[trim=2cm 0.5cm 2cm 0.5cm,width=0.46\linewidth]{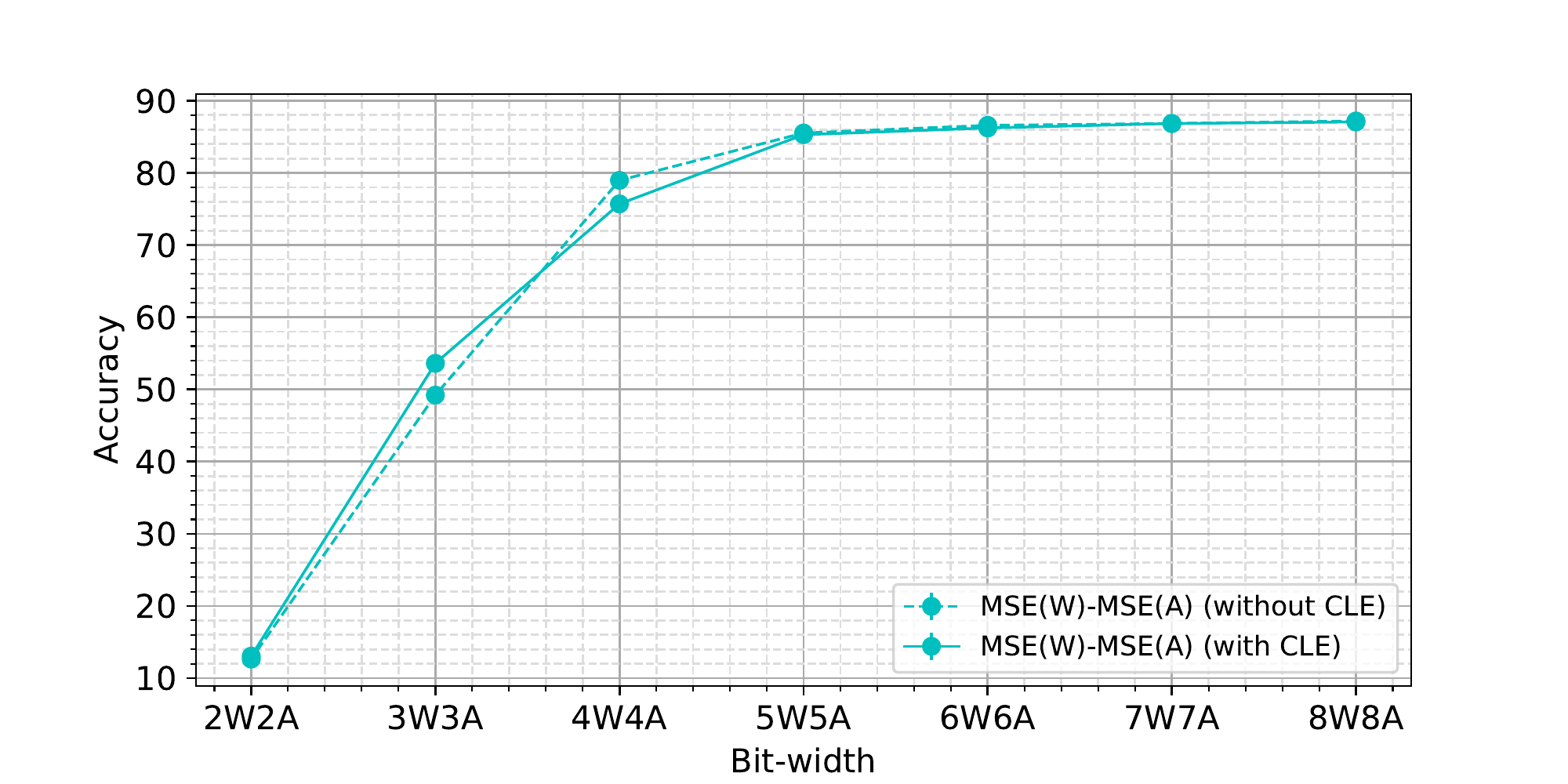}} 
    & 
    \subcaptionbox{\label{2st-fig}Pre-processing: DS-CNN(Speech Command)}{\includegraphics[trim=2cm 0.5cm 2cm 0.5cm,width=0.46\linewidth]{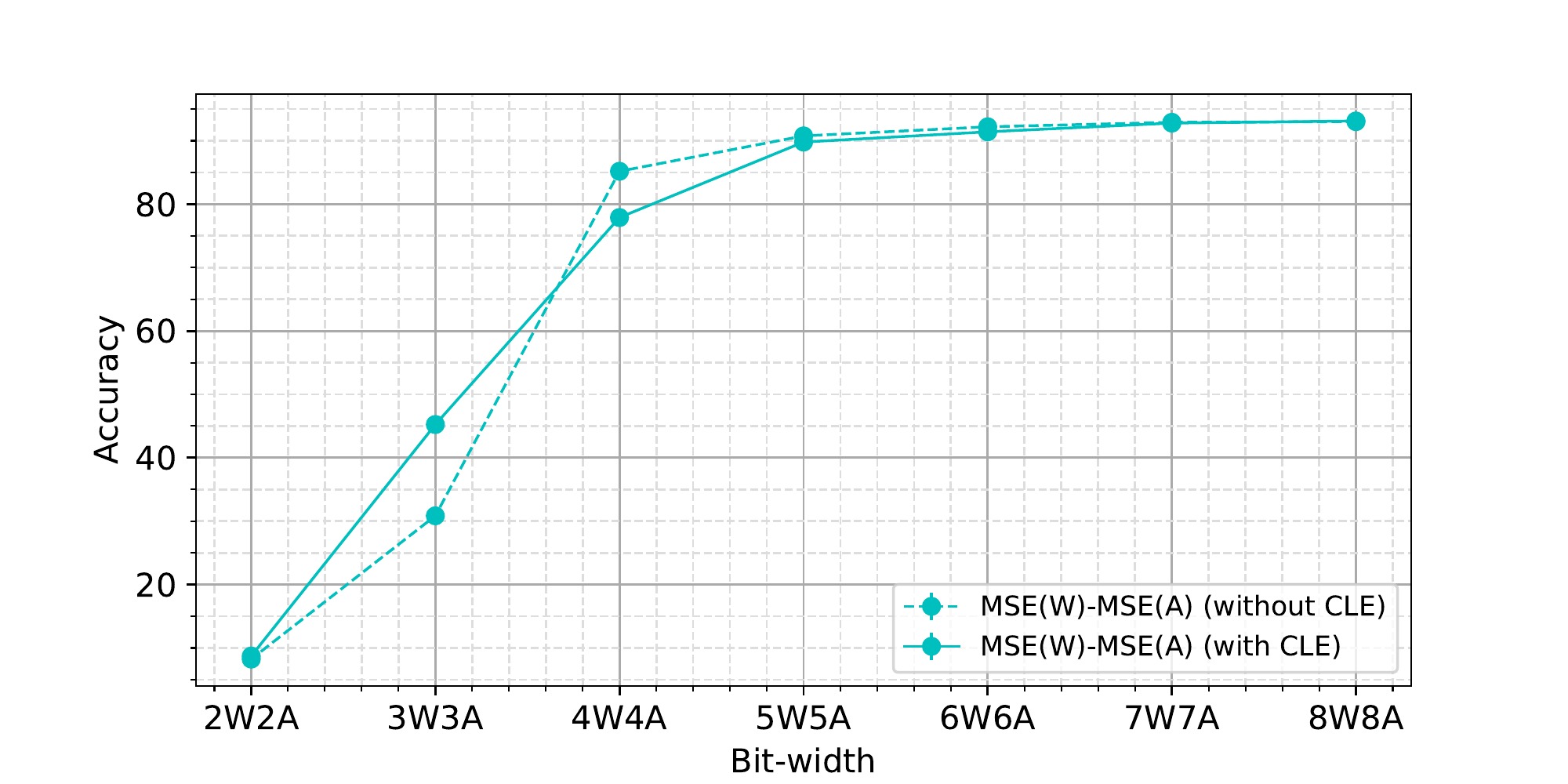}} 
  \end{tabular}
  \caption{Ablation study on FP model preprocessing and quantizer initialization for both weights and activations. (a) and (b) show the accuracy after quantizer intialization. (c) and (d) show the accuracy after MSE-bsed quantizer initialization with and without CLE as the FP model preprocessing.}
  \label{fig.abla_init}
\end{figure}
\begin{figure}[htb]
  \centering
  \begin{tabular}{cc}
    \subcaptionbox{\label{1st-fig}Res8 (CIFAR10)}{\includegraphics[trim=2cm 0.5cm 2cm 0.5cm, width=0.46\linewidth]{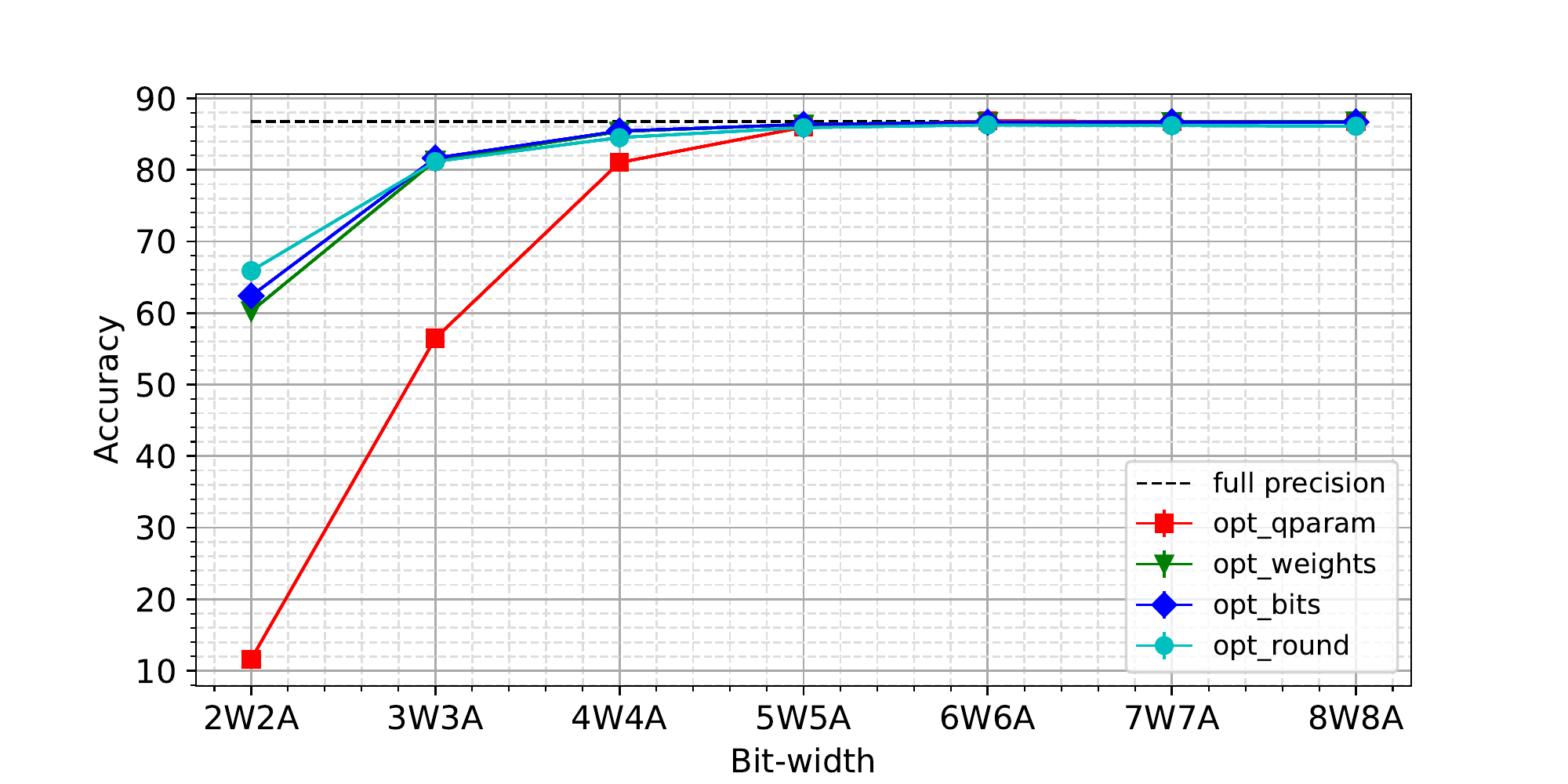}} 
    & 
    \subcaptionbox{\label{2st-fig}DS-CNN(Speech Command)}{\includegraphics[trim=2cm 0.5cm 2cm 0.5cm, width=0.46\linewidth]{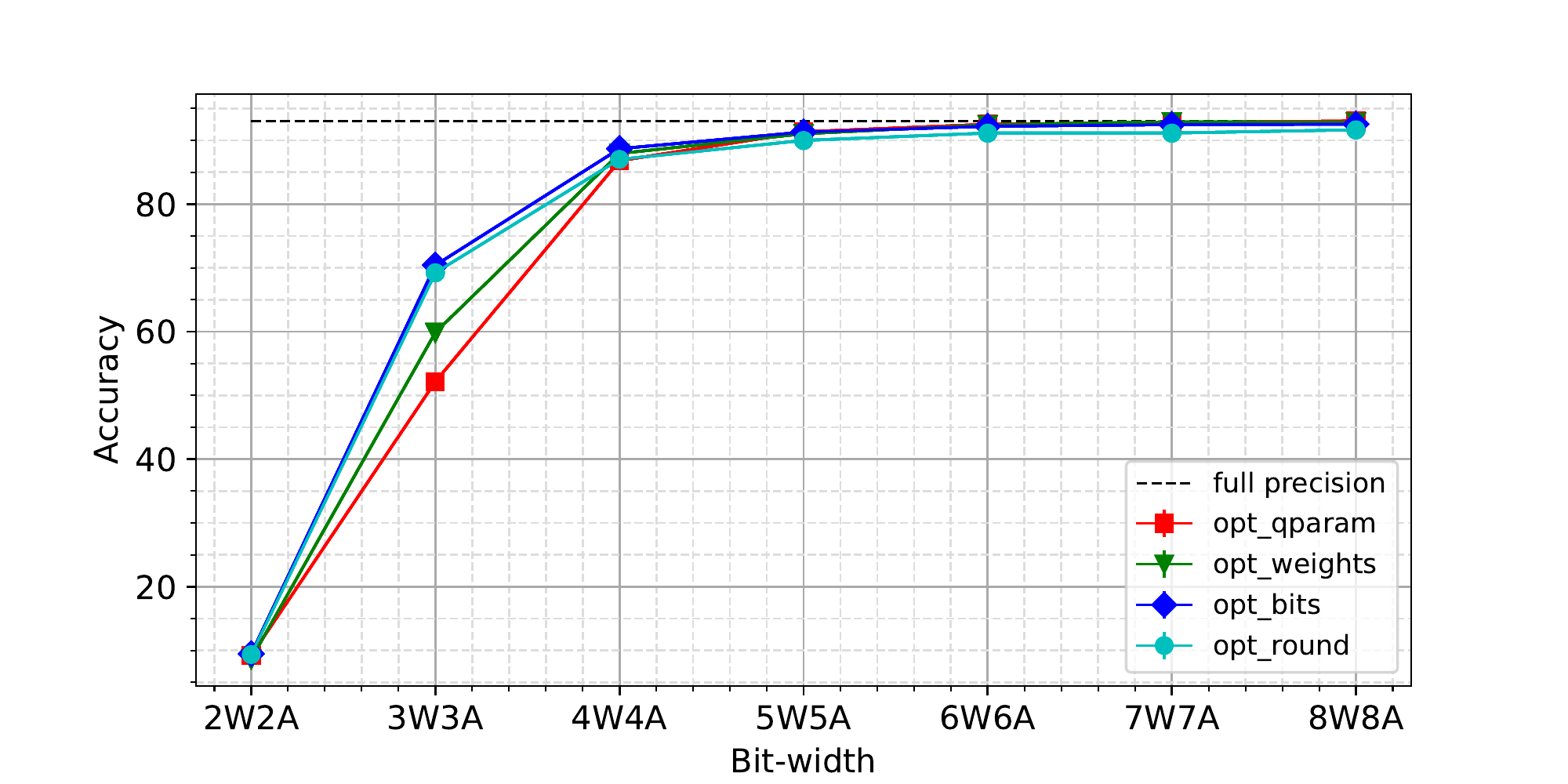}} 
  \end{tabular}
  \caption{Ablation study on optimization variables during quantization optimization. Better zoom in to view the results.}
  \label{fig.abla_opt_var}
\end{figure}
\textit{FP model preprocessing and quantizer initialization.} We consider FP model pre-processing and quantizer initialization together in this study to see how the combination of the two affects the accuracy. Figure~\ref{fig.abla_init} (a) and (b) show the comparison of model accuracy with different initialization methods. We need to ignore the results here for $2W2A$ because the quantized model after initialization performs as a random classifier caused by large quantization noise. From the results, we can see that MSE-based quantizer initialization usually outperforms MinMax-based one. And MSE-based initialization for both weights and activations performs the best consistently, this is aligned with previous study on quantizer initialization~\cite{nagel2021}. For FP model preprocessing, we consider cross layer equalization (CLE) because it is effective without changing the architecture of the model. Figure~\ref{fig.abla_init} (c) and (d) compare the model accuracy with and without CLE after MSE-based initialization. We see that there is no accuracy discrepancy for higher bitwidth. Although we use per-channel quantization for weights, CLE still helps to improve the accuracy for low precision cases ($3W3A$), while it has negative impact for the precision in between($4W4A$ or $5W5A$). This is because activation tensors have per-tensor quantization and CLE also changes the distribution of activation tensors.

\textit{Optimization variable.} As discussed in section \ref{sec.ptq_pipeline}, different variables can be optimized to minimize the layer-wise loss. We investigate the effect of optimizing different variables. Please note that only the weights are quantized while activation are in full precision for this study. As shown in Figure~\ref{fig.abla_opt_var}, we can see that \textit{opt\_qparam} is clearly less effective compared with the other three, especially when the bit-width is low. It is because that only the quantization parameters (zero-points and offsets) are trainable, which restricted its capability to to compensate for the large quantization noise introduced by low precision quantization. \textit{opt\_round} and \textit{opt\_bits} usually outperforms \textit{opt\_weights}. We believe that the STE~\cite{bengio2013ste} used in \textit{opt\_weights} causes biased gradient estimation under low-precision cases that affects the efficacy of optimization. And the constrained optimization space in \textit{opt\_bits} and \textit{opt\_round} makes them less prone to over-fitting with limited calibration data. 

\begin{figure}[htb]
  \centering
  \begin{tabular}{cc}
    \subcaptionbox{\label{1st-fig}Res8 (CIFAR10)}{\includegraphics[trim=2cm 0.5cm 2cm 0.5cm, width=0.46\linewidth]{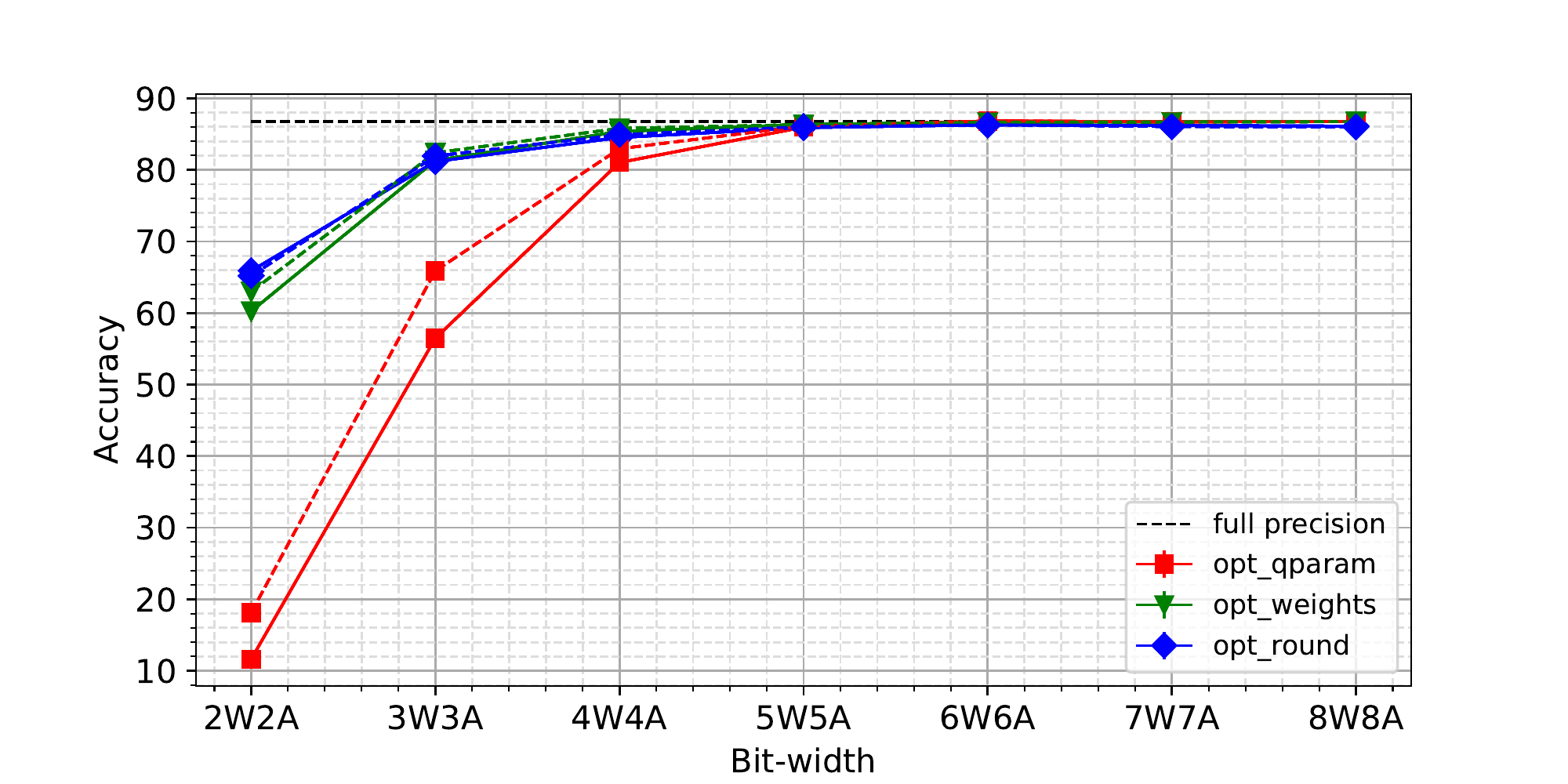}} 
    &
    \subcaptionbox{\label{2st-fig}DS-CNN(Speech Command)}{\includegraphics[trim=2cm 0.5cm 2cm 0.5cm, width=0.46\linewidth]{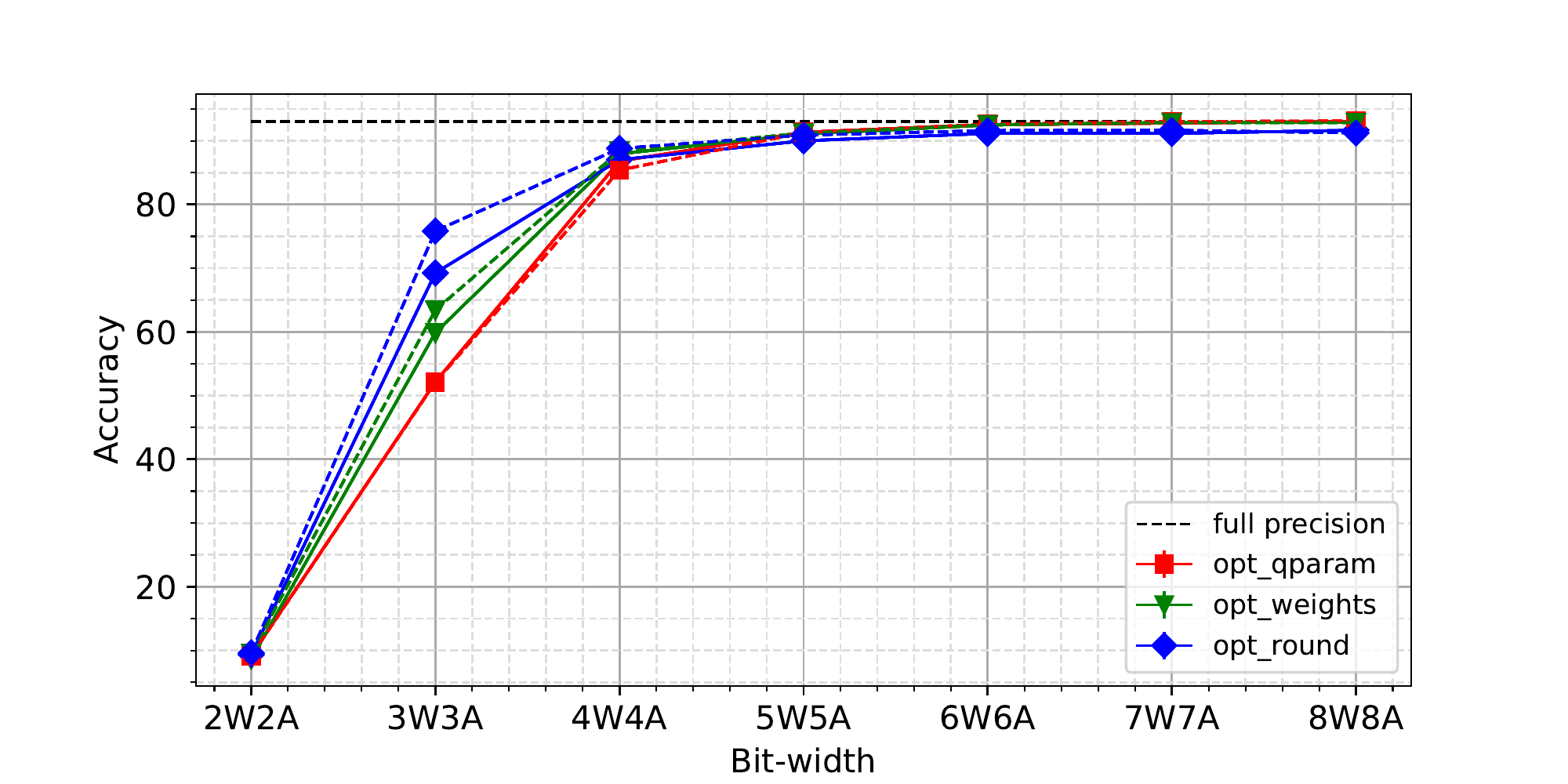}}
    \end{tabular}
  \caption{Ablation study on optimization granularity during quantization optimization. The solid lines represent the layerwise optimization and the dotted lines represent the blockwise optimization. Better zoom in to view the results.}
  \label{fig.abla_opt_granu}
\end{figure}

\textit{Optimization granularity.} We conduct the study on optimization granularity and the results are shown in Figure \ref{fig.abla_opt_granu}. The solid lines represent the layerwise optimization and the dotted lines represent the blockwise optimization. \textit{opt\_bits} is excluded here because it doesn't support blockwise optimization by design. The block is defined as the basic residual block in Res8 and the depth separable convolution in DS-CNN. We observe that both layerwise optimization and blockwise optimization have very similar performance when the models are quantized to $4W4A$ and higher, regardless of the different optimization variables. For quantization with precision lower than $4W4A$ , blockwise optimization provides quite significant improvement on the accuracy for Res8 quantized to $3W3A$ and $2W2A$, and for DS-CNN quantized to $3W3A$.


\begin{figure}[htb]
  \centering
  \begin{tabular}{cc}
    \subcaptionbox{Res8 (CIFAR10)}{\includegraphics[trim=2cm 0.5cm 2cm 0.5cm, width=0.45\linewidth]{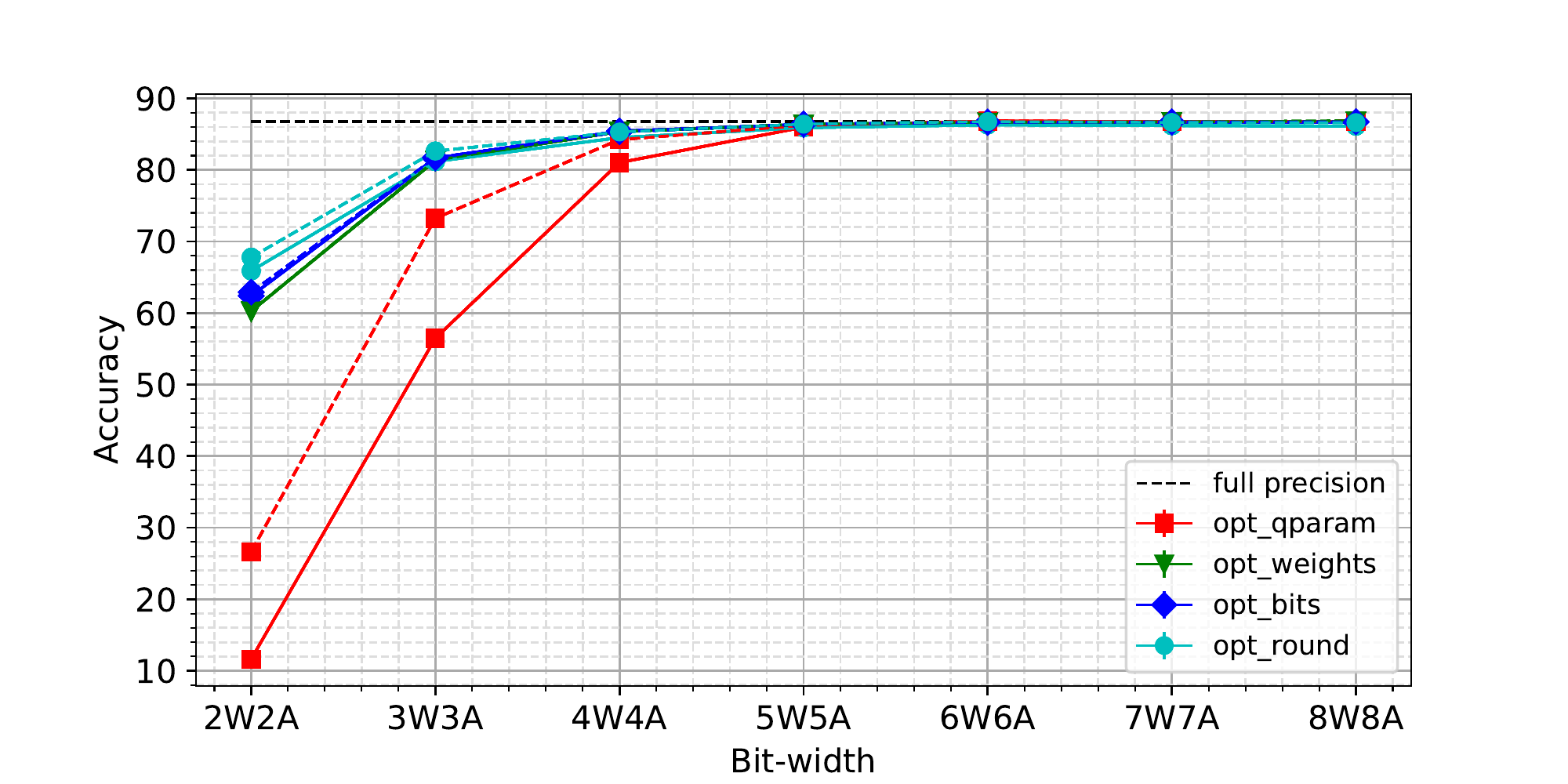}} 
    &
    \subcaptionbox{DS-CNN(Speech Command)}{\includegraphics[trim=2cm 0.5cm 2cm 0.5cm, width=0.45\linewidth]{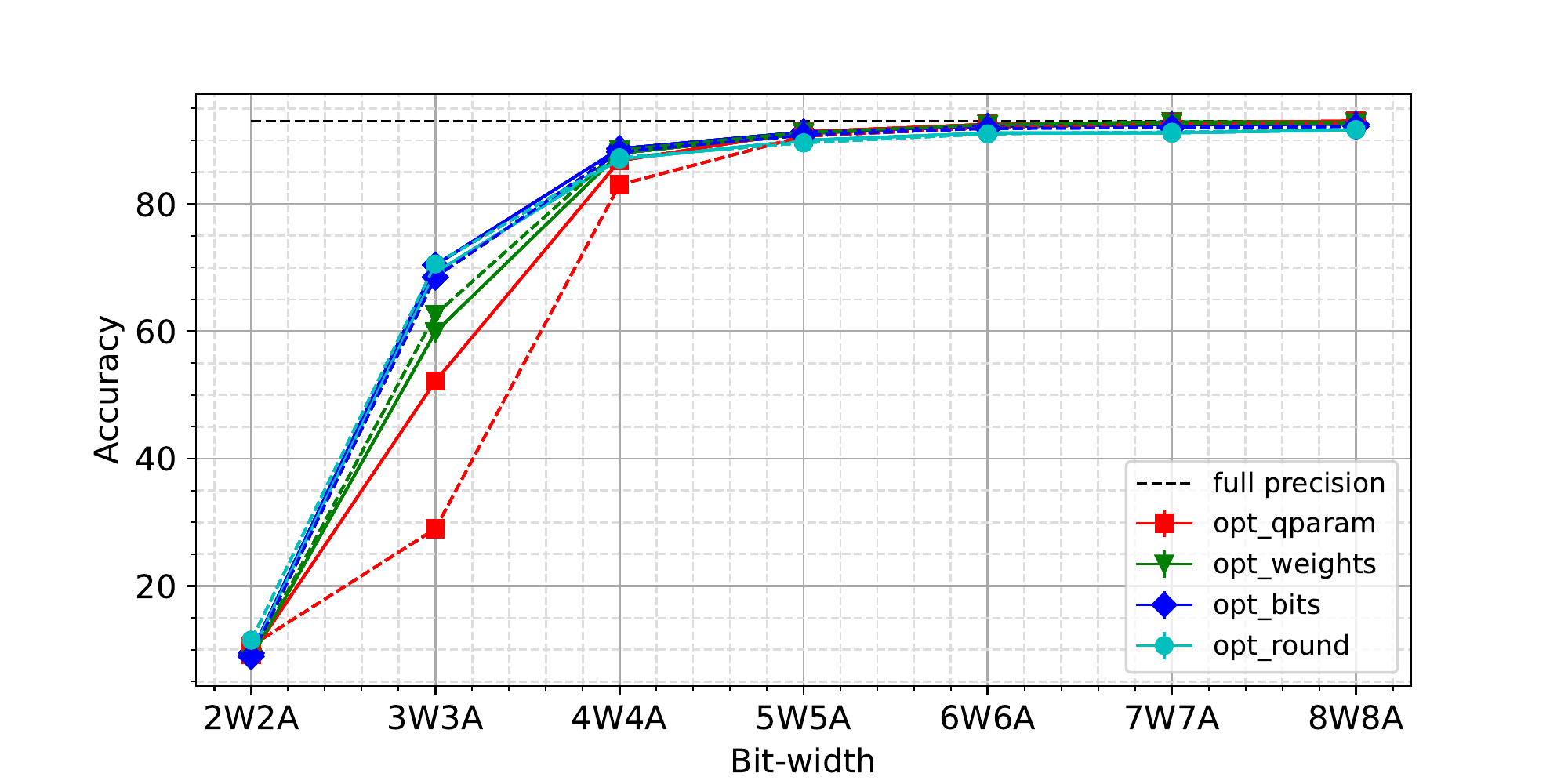}}
  \end{tabular}
  \caption{Ablation study on bias tuning for LP model post-processing. The solid lines represent optimization without post-processing and the dotted lines represent the optimization with post-processing. Better zoom in to view the results.}
  \label{fig.abla_post}
\end{figure}

\textit{LP model Post-processing.} We also conduct experiment to show the impact of bias tuning as the low precision model post-processing. As show in Figure \ref{fig.abla_post}, bias tuning consistently improve the accuracy for all the methods (except for the case of DS-CNN using \textit{opt\_qparam}). For Res8 using \textit{opt\_qparam}, it has more than $10\%$ improvement for both $3W3A$ and $2W2A$  quantization. This empirically shows the effectiveness of end-to-end bias tuning.

\subsection{Results}
\begin{table}[t]
\centering
\begin{scriptsize}
\begin{tabular}{cccccc}
\hspace*{0.5cm} & \hspace*{1cm} & \hspace*{1cm} & \hspace*{1cm}\\[-1em]
\hline
\textbf{Bitwidth} & \multirow{2}*{\textbf{Method}}  & \textbf{Res8} & \textbf{DS-CNN} & \textbf{MobileNetv1} & \textbf{CNN} \\
\textbf{(W/A)} &   & \textbf{(CIFAR10)} & \textbf{(KWS)} & \textbf{(VWW)} & \textbf{(UCI-HAR)} \\
\hline \hline
\multirow{1}*{32/32} & full precision & 86.75 & 93.03 & 85.10 & 90.36\\
\hline
\multirow{4}*{8/8} & opt\_qparam & \textbf{86.81\textpm0.06}  & \textbf{92.63\textpm0.14} & 85.13\textpm0.12 & \textbf{90.39\textpm0.17} \\ 
& opt\_weights & 86.76\textpm0.04 & 92.55\textpm0.06 & 85.14\textpm0.05 & 90.81\textpm0.24  \\ 
& opt\_bits  & 86.70\textpm0.10   & 91.93\textpm0.10 & \textbf{85.28\textpm0.15} & 90.04\textpm0.13 \\ 
& opt\_round & 86.65\textpm0.14   & 91.65\textpm0.20 & 84.98\textpm0.21 & 90.37\textpm0.26 \\
\hline

\multirow{4}*{7/7} & opt\_qparam & \textbf{86.79\textpm0.11} & 92.57\textpm0.10 & \textbf{85.18\textpm0.16} & 90.54\textpm0.24  \\ 
& opt\_weights & 86.69\textpm0.08 & \textbf{92.62\textpm0.14}  & 85.15\textpm0.17 & \textbf{90.88\textpm0.31}  \\ 
& opt\_bits & 86.64\textpm0.08    & 91.92\textpm0.15  & 85.16\textpm0.17 & 89.98\textpm0.15 \\ 
& opt\_round & 86.70\textpm0.14   & 91.54\textpm0.21  & 84.87\textpm0.11 & 90.37\textpm0.35 \\
\hline
    
\multirow{4}*{6/6} & opt\_qparam & 86.67\textpm0.15 & 92.01\textpm0.19 & 84.90\textpm0.14 & 90.24\textpm0.11 \\ 
& opt\_weights & \textbf{86.71\textpm0.14}  & \textbf{92.19\textpm0.19} & 85.06\textpm0.10 & \textbf{90.86\textpm0.22} \\ 
& opt\_bits    & 86.63\textpm0.06  & 91.80\textpm0.14 & \textbf{85.11\textpm0.10} & 90.03\textpm0.31 \\ 
& opt\_round & 86.64\textpm0.06    & 90.56\textpm0.31 & 84.81\textpm0.18 & 90.20\textpm0.23 \\ 
\hline
\multirow{4}*{5/5} & opt\_qparam & 86.35\textpm0.03 & \textbf{90.86\textpm0.28} & 84.33\textpm0.19 & 90.61\textpm0.41 \\ 
& opt\_weights & 86.47\textpm0.12  & 90.82\textpm0.20 & \textbf{84.82\textpm0.15} &  \textbf{91.00\textpm0.47}  \\ 
& opt\_bits    & 86.37\textpm0.15  & 90.42\textpm0.36 & 84.39\textpm0.16 & 90.09\textpm0.21 \\ 
& opt\_round & \textbf{86.64\textpm0.10}    & 90.13\textpm0.93 & 84.21\textpm0.24 & 90.41\textpm0.30 \\ 
\hline
\multirow{4}*{4/4} & opt\_qparam & 84.61\textpm0.15 & 83.31\textpm1.02 & 78.40\textpm1.52 & 91.12\textpm0.22 \\ 
& opt\_weights & \textbf{85.91\textpm0.18} & \textbf{88.41\textpm0.25}  & \textbf{82.96\textpm0.23} & \textbf{90.76\textpm0.13} \\ 
& opt\_bits    & 85.33\textpm0.09 & 87.88\textpm0.36  & 81.72\textpm0.62& 90.38\textpm0.32 \\ 
& opt\_round & 85.34\textpm0.32   & 88.33\textpm0.95  & 81.89\textpm0.50& 90.57\textpm0.35 \\ 
\hline
\multirow{4}*{3/3} & opt\_qparam & 74.30\textpm0.53 & 29.58\textpm4.60 & 56.79\textpm1.43 & 90.57\textpm0.23 \\ 
& opt\_weights & 82.62\textpm0.13  & 63.99\textpm2.98 & 76.03\textpm1.36 & 90.83\textpm0.31 \\ 
& opt\_bits    & 81.81\textpm0.14  & 63.95\textpm2.66 & 74.45\textpm0.71 & 90.32\textpm0.27 \\ 
& opt\_round   & \textbf{83.21\textpm0.41}  & \textbf{75.22\textpm1.57} & \textbf{76.71\textpm1.21} & \textbf{91.38\textpm1.15} \\ 
\hline
\multirow{4}*{2/2} & opt\_qparam & 25.26\textpm2.2 & 9.16\textpm1.18 & 47.15\textpm0.00 & 82.71\textpm0.76 \\ 
& opt\_weights & 62.81\textpm0.53  & 8.83\textpm0.85 & 62.24\textpm0.58 & \textbf{90.46\textpm0.54} \\ 
& opt\_bits    & 63.72\textpm0.47  & 8.77\textpm0.23 & \textbf{62.48\textpm0.34} & 86.74\textpm0.59 \\ 
& opt\_round   & \textbf{69.40\textpm0.74}  & \textbf{9.42\textpm0.51} & 60.58\textpm0.96 & 89.63\textpm0.83\\ 
\hline

\end{tabular}
\end{scriptsize}

\caption{Comparison of PTQ pipelines with different optimization variables while fixing other options the best in the pipeline. }
\label{tab:e2e}
\end{table}
To compare end-to-end performance with both weight and activation quantization enabled, we fix all the options in the pipeline using the best ones based on ablation study results, except for the optimization variables. We use symmetric per-channel quantization for weights and asymmetric per-tensor quantization for activations. The FP model pre-processing is enabled and MSE-based initialization is applied on both weight quantizers and activation quantizers. And we use layerwise optimization with bias tuning enabled as LP model post-processing. The results are shown in Table \ref{tab:e2e}. For a higher bit-width ($5W5A$ to $8W8A$), different PTQ pipelines perform similarly with very small accuracy drop compared to the full precision accuracy. However, for a lower bit-width, different pipelines show big performance discrepancy. The pipelines using \textit{opt\_round} usually perform better. However, even for the best quantization algorithm, there is still big accuracy drop when going to ultra-low precision. The accuracy for $2W2A$ is $69.40\%$ for CIFAR10, with accuracy drop as large as $17.5\%$. 

\begin{figure}[ht]
  \centering
  \begin{tabular}{c}
    \subcaptionbox{}{\includegraphics[trim=2cm 0.25cm 2cm 0.5cm, width=0.9\linewidth]{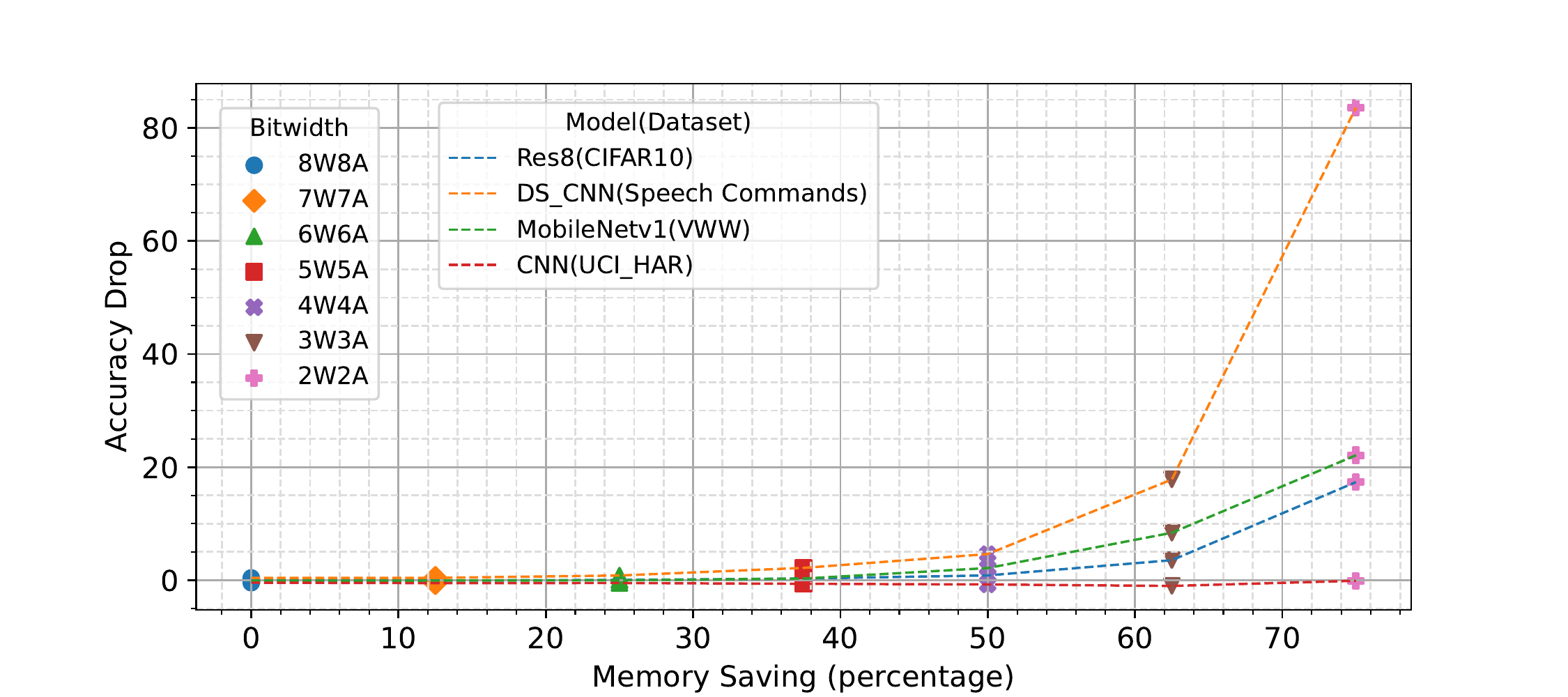}} 
    \\
    \subcaptionbox{}{\includegraphics[trim=2cm 0.25cm 2cm 0.5cm, width=0.9\linewidth]{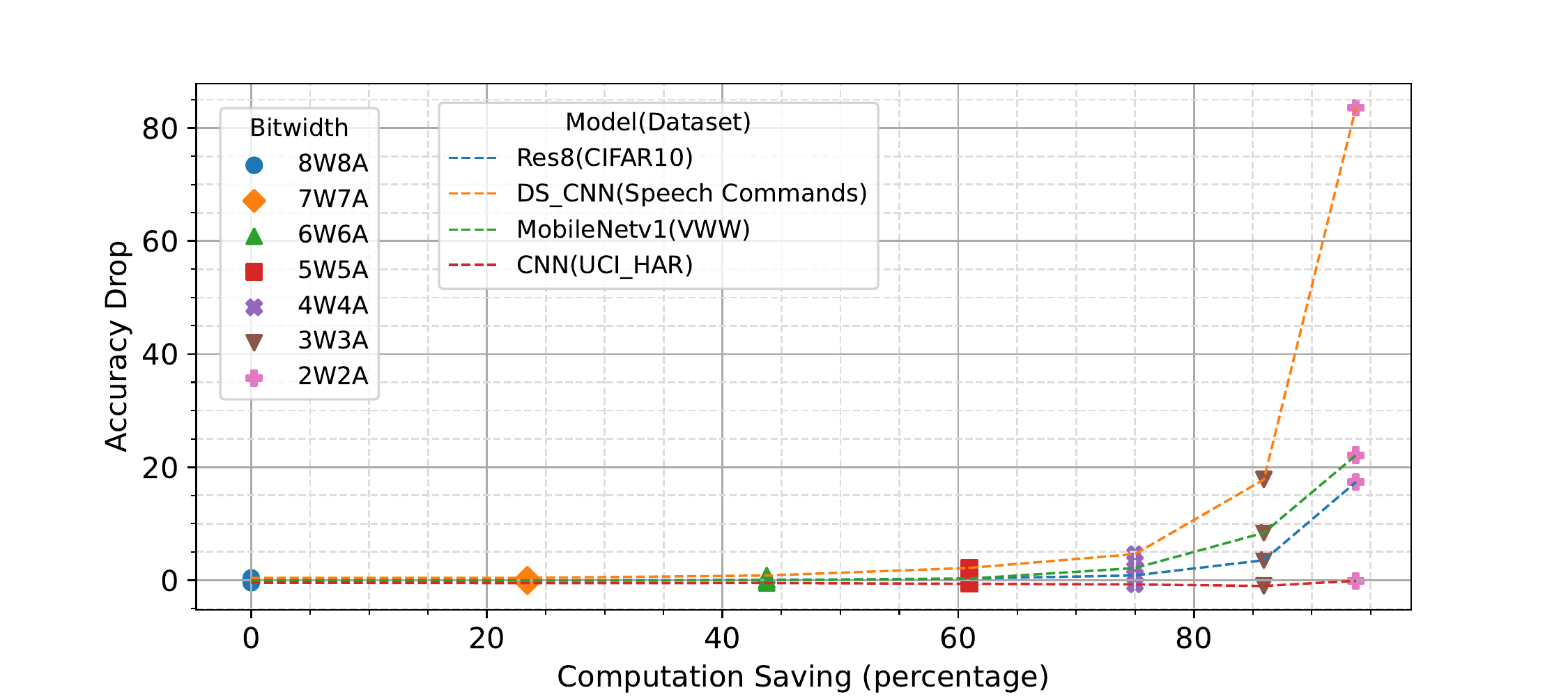}}
    \end{tabular}
  \caption{Tradoff between accuracy and memory/computation saving. $4W4A$ is a sweet point having large memory/computation saving with small accuracy drop. Some models can go lower precision with negligible accuracy drop.}
  \label{fig.saving_view}
\end{figure}
To better show the benefits of low precision quantization, we view the results on two axes: the accuracy drop of the best PTQ pipeline, and corresponding computation saving (or memory saving) with different bit-widths. We use bit operation (BOP) as a system-agnostic proxy to the computation complexity. The BOP count of a layer $l$ is computed as: $BOP(l)=MACs(l)b_wb_a$, where $MACs(l)$ is the multiply-accumulate count of the layer, $b_w$ is the bit-width of the weight and $b_a$ is the bit-width of the input. We use peak memory as the measurement of the memory consumption, which is computed as $peak\_memory=total\_weight*b_w+peak\_activation*b_a$. it measures the maximum memory required to store the weights and activations during inferences, assuming that previous intermediate activations can be effectively released if they are not used by the later layers. The results are shown in Figure~\ref{fig.saving_view}. We can see that $4W4A$ is a sweet point that has $50\%$ memory saving and $75\%$ computation saving with small accuracy drop ($<5\%$). This is mismatching with most inference frameworks that support only 8-bit quantization~\cite{ghamari2021}. Some models can even tolerant lower precision quantization with negligible accuracy drop. For example, the CNN model for UCI-HAR can be quantized to 2W2A with less than $2\%$ accuracy drop. However, this could be a model/task dependent behavior. For most of the models, we have to sacrifice the accuracy if going for ultra-low precision such as 3W3A or 2W2A, and thus facing the accuracy and memory/computation trade-off. We can clearly see the benefits of low precision quantization in terms of computation and memory saving (and eventually power saving). 

\section{Conclusion and Future Directions}
\label{sec:conclu}
In this paper, with a unified quantization pipeline, we benchmark PTQ algorithms on carefully selected models for tinyML. We conduct ablations to reveal the key design choices of the pipeline. We show that low precision quantization with the best PTQ algorithm can have huge memory and computation savings with negligible accuracy drop, although there is still big accuracy drop if quantized models to ultra-low precision. Our empirical study provides useful data points on low precision quantization and points out potential future directions for improvement on algorithm design and system support for tinyML.

How to further improve the accuracy for ultra-low precision quantization is an important direction for future research. Based on our unified PTQ pipeline, new improvement on different components can further reduce the accuracy degradation. And mix-precision quantization and quantization aware training could be potential solutions. Besides, existing embedded AI frameworks, such as TensorFlow Lite Micro~\cite{tflitemicro2021} and STM32Cube.AI~\cite{xcubeai2021}, can only support computation in floating point or 8-bit integers. The framework with lower precision quantization support is not widely available. It first leaves big room to squeeze the model by combining quantization with other model compression techniques. And it is important to investigate the best combination in practice. At the meanwhile, developing efficient embedded AI frameworks with low precision quantization support will be particularly useful for deployment of tinyML models.


\bibliographystyle{ACM-Reference-Format}
\bibliography{main}

\appendix
\section{Model architectures}
Here we show the basic blocks used to define the models and the detailed architecture of each TinyML model. 
\begin{figure}[!htb]
    \centering
    \begin{tabular}{ccc}
    \subcaptionbox{{\scriptsize DSConv}}{\includegraphics[width=0.20\linewidth]{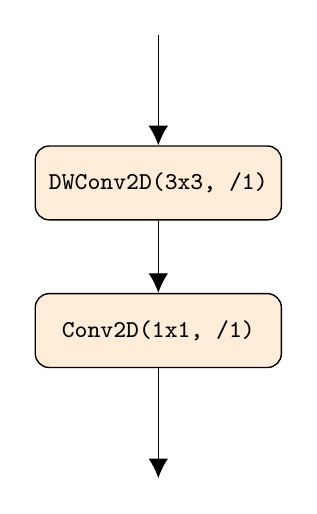}} 
    &
    \subcaptionbox{{\scriptsize ResBlock(stride=$1$)}}{\includegraphics[ width=0.27\linewidth]{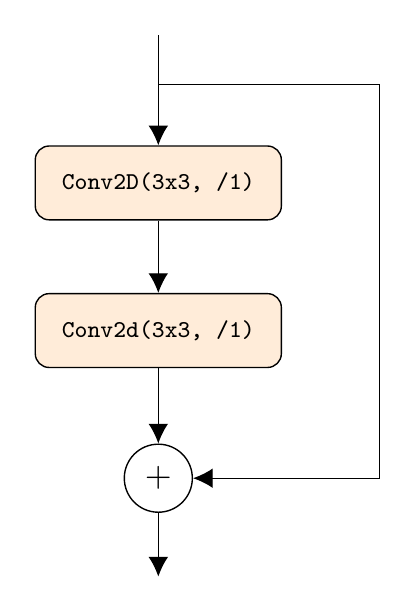}} 
    &
    \subcaptionbox{{\scriptsize ResBlock(stride=$2$)}}{\includegraphics[ width=0.39\linewidth]{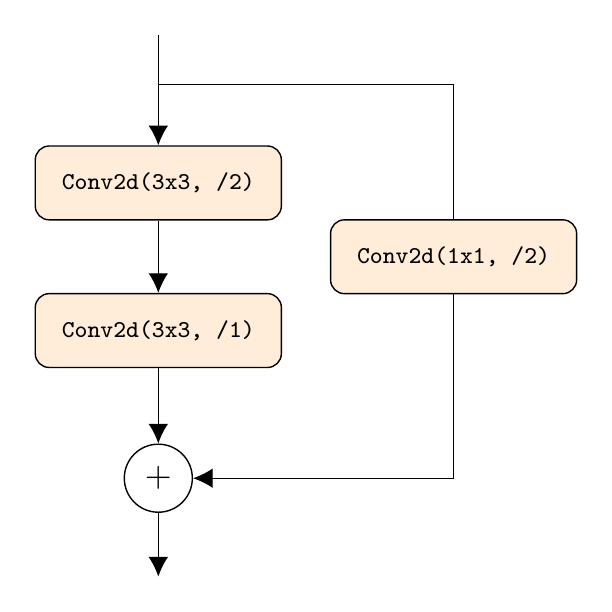}} 
    \end{tabular}
    \caption{The basic blocks used in the model architectures.}
    \label{fig:blocks}
\end{figure}


\begin{table}[!htb]
\centering
\begin{tiny}
\begin{tabular}{|c|c|c|c|}
\hline
\textbf{Input} & \textbf{Operator} & \textbf{\#out} & \textbf{stride}\\ 
\hline
$32\times32\times3$ & Conv2d, $3\times3$ & 16 & 1 \\ 
\hline
$32\times32\times16$ & ResBlock & 16 & 1 \\ 
\hline
$32\times32\times16$ & ResBlock & 32 & 2 \\ 
\hline
$16\times16\times32$ & ResBlock & 64 & 2 \\ 
\hline
$8\times8\times64$ & AvgPool, $8\times8$ & - & 1 \\ 
\hline
$1\times64$ & Fully Connected & 10 & - \\
\hline
\end{tabular}
\end{tiny}
\caption{Res8 (CIFAR10) architecture definition}
\end{table}

\begin{table}[h]
\centering
\begin{tiny}
\begin{tabular}{|c|c|c|c|}
\hline
\textbf{Input} & \textbf{Operator} & \textbf{\#out} & \textbf{stride}\\ \hline
$10\times49\times1$ & Conv2d, $4\times10$ & 64 & 2 \\ \hline
$5\times25\times64$ & DSConv & 64 & 1 \\ \hline
$5\times25\times64$ & DSConv & 64 & 1 \\ \hline
$5\times25\times64$ & DSConv & 64 & 1 \\ \hline
$5\times25\times64$ & DSConv & 64 & 1 \\ \hline
$5\times25\times64$ & AvgPool, 5x25 & - & 1 \\ \hline
$1\times64$ & Fully Connected & 12 & - \\ \hline
\end{tabular}
\end{tiny}
\caption{DSCNN(Speech Commands) architecture definition}
\end{table}

\begin{table}[h]
\centering
\begin{tiny}
\begin{tabular}{|c|c|c|c|}
\hline
\textbf{Input} & \textbf{Operator} & \textbf{\#out} & \textbf{stride}\\ \hline
$96\times96\times3$ & Conv2d, $3\times3$ & 8 & 2 \\ \hline
$48\times48\times8$ & DSConv & 16 & 1 \\ \hline
$48\times48\times16$ & DSConv & 32 & 2 \\ \hline
$24\times24\times32$ & DSConv & 32 & 1 \\ \hline
$24\times24\times32$ & DSConv & 64 & 2 \\ \hline
$12\times12\times64$ & DSConv & 64 & 1 \\ \hline
$12\times12\times64$ & DSConv & 128 & 2 \\ \hline
$6\times6\times128$ & DSConv & 128 & 1 \\ \hline
$6\times6\times128$ & DSConv & 128 & 1 \\ \hline
$6\times6\times128$ & DSConv & 128 & 1 \\ \hline
$6\times6\times128$ & DSConv & 128 & 1 \\ \hline
$6\times6\times128$ & DSConv & 256 & 2 \\ \hline
$3\times3\times256$ & DSConv & 256 & 1 \\ \hline
$3\times3\times256$ & AvgPool, $3\times3$ & - & 1 \\ \hline
$1\times256$ & Fully Connected & 2 & - \\ \hline
\end{tabular}
\end{tiny}
\caption{MobileNetv1(VWW) architecture definition}
\end{table}

\begin{table}[!htb]
\centering
\begin{tiny}
\begin{tabular}{|c|c|c|c|}
\hline
\textbf{Input} & \textbf{Operator} & \textbf{\#out} & \textbf{stride}\\ 
\hline
$128\times9$ & Conv1d, $3$ & 64 & 1 \\ 
\hline
$126\times64\times16$ & Conv1d, $3$ & 64 & 1 \\ 
\hline
$124\times64$ & MaxPool & - & 2 \\ 
\hline
$62\times 64$ & Flatten & - & - \\ 
\hline
$1 \times 3960$ & Fully Connected & 6 & - \\ 
\hline
\end{tabular}
\end{tiny}
\caption{CNN(UCI-HAR) architecture definition}
\end{table}

\end{document}